\documentclass[preprint,11pt,authoryear]{elsarticle}

\usepackage{amssymb}
\usepackage{amsmath}
\usepackage{setspace}
\usepackage{amsthm}
\usepackage{color}

\usepackage{algpseudocode}
\usepackage{lscape}

\usepackage{pgfplots}
\pgfplotsset{compat=1.12}

\usepackage{graphics}
\usepackage{graphicx}

\usepackage{caption}
\usepackage{subcaption}
\usepackage{float}

\usepackage[boxed, algoruled, linesnumbered, inoutnumbered]{algorithm2e}
\usepackage{mathrsfs}
\usepackage{multicol}
\usepackage{multirow}
\usepackage{empheq}

\usepackage{booktabs}
\usepackage{hyperref}
\usepackage{tabularx}
\usepackage{adjustbox}
\usepackage{xcolor}
\usepackage{mathtools}
\usepackage{threeparttable}

\usepackage{tikz}
\usetikzlibrary{calc,arrows.meta}
\tikzset{%
  >={Latex[width=2mm,length=2mm]},
            base/.style = {rectangle, rounded corners, draw=black,
                           minimum width=4cm, minimum height=1cm,
                           text centered, font=\sffamily},
  activityStarts/.style = {base, fill=blue!30},
       startstop/.style = {base, fill=red!30},
    activityRuns/.style = {base, fill=green!30},
         process/.style = {base, minimum width=2.5cm, fill=orange!15,
                           font=\ttfamily},
}

\DeclareMathOperator*{\argmin}{arg\!min}

\usepackage[utf8]{inputenc} 
\usepackage[T1]{fontenc} 
\usepackage{paralist}

\biboptions{numbers,sort}
\newcounter{magicrownumbers}

\journal{NEUNET}

\begin{document}

\linespread{1.25}

\begin{frontmatter}

\title{A Population-based Hybrid Approach to Hyperparameter Optimization for Neural Networks}

\author[CEFET/RJ]{Marcello Serqueira}\ead{marcello.serqueira@eic.cefet-rj.br}
\author[CEFET/RJ]{Pedro Henrique Gonz\'{a}lez}\ead{pegonzalez@eic.cefet-rj.br}
\author[CEFET/RJ]{Eduardo Bezerra\corref{cor1}}\ead{ebezerra@cefet-rj.br}

\cortext[cor1]{Corresponding author}

\address[CEFET/RJ]{Graduate Program in Computer Science\\ Federal Center for Technological Education of Rio de Janeiro (CEFET/RJ)\\ Rio de Janeiro, Brazil}

\begin{abstract}
In recent years, large amounts of data have been generated, and computer power has kept growing. This scenario has led to a resurgence in the interest in artificial neural networks. One of the main challenges in training effective neural network models is finding the right combination of hyperparameters to be used. Indeed, the choice of an adequate approach to search the hyperparameter space directly influences the accuracy of the resulting neural network model. Common approaches for hyperparameter optimization are Grid Search, Random Search, and Bayesian Optimization. There are also population-based methods such as CMA-ES. In this paper, we present HBRKGA, a new population-based approach for hyperparameter optimization. HBRKGA is a hybrid approach that combines the Biased Random Key Genetic Algorithm with a Random Walk technique to search the hyperparameter space efficiently. Several computational experiments on eight different datasets were performed to assess the effectiveness of the proposed approach. Results showed that HBRKGA could find hyperparameter configurations that outperformed (in terms of predictive quality) the baseline methods in six out of eight datasets while showing a reasonable execution time.
\end{abstract}


\end{frontmatter}


\section{Introduction}
\label{sec:introduction}

Artificial Neural Networks (ANN) are an old approach to Machine Learning that have witnessed a renewed interest both from industry and academia in recent years \citep{goodfellow2016deep}. This interest is motivated by cases of success in several application domains, such as audio recognition, image recognition, and language translation. A particular advancement in the field of ANN in the last decade is related to the fact that the research community has been gradually learning to deal with the engineering problem of training neural networks comprised of several hidden layers. This renaissance of neural networks has been called Deep Learning \citep{lecun2015deep}.

A prerequisite to training a neural network model is to come up with a particular combination of values of hyperparameters, such as the number of hidden layers, the number of artificial neurons in each, the learning rate, the activation functions to be used, to name a few. Only after a particular set of hyperparameters has been chosen can the training process tune the parameters (i.e., the weights) of the ANN. Many hyperparameters have continuous domains, which accelerate the exponential growth of possible values combination. The huge multidimensional space resulting from combinations of several hyperparameters is an even more significant challenge in the Deep Learning era. So much that the area of automatic machine learning (AutoML) has emerged to study automation and optimization of machine learning models. AutoML aims to join and to automate the whole process of machine learning (hyperparameter optimization, architectures, optimization algorithms) for the creation of accurate models without the need for deep statistical knowledge and programming \citep{he2019automl}.

A critical aspect of AutoML is hyperparameter optimization. There are several techniques to perform a search in the hyperparameter space. In general, these techniques work as a procedure that runs an outer loop in the learning process: this procedure suggests a combination of hyperparameter values, which are then used to optimize the set of parameters in the neural network. These techniques usually do not present any assumption for performing the search; only a fixed range of values is defined by the user to be explored.

Two popular approaches to optimize hyperparameters are Grid Search and Random Search. \citet{bergstra2012random} performed Random Search experiments in comparison to the results of the experiments obtained by \citet{larochelle2007empirical}. They showed that in most datasets, Random Search was able to overcome Grid Search, both in accuracy and in computational performance. Thus far, Random Search has shown to be an efficient alternative to the Grid Search optimization strategy.

One downside of both Grid Search and Random Search is that they do not try to improve based on previously tested hyperparameters combinations. Hence, in recent years more intelligent methods have been explored to perform this optimization. Among the most covered is Bayesian Optimization~\citep{snoek2012practical}. The method is different from Random and Grid Search because it allows for ANN's optimization without the need to define the search space manually with high precision. Unfortunately, Bayesian Optimization is computationally expensive, since its time complexity is cubic on the number of samples seen before~ \citep{snoek2015scalable}.

A recent alternative to hyperparameter tuning is the family of population-based methods~\citep{simon2013evolutionary}. These are evolutionary algorithms that aim to evolve individuals in a hyperparameter configuration population by applying operations such as crossover and mutation~\citep{hutter2019automated}. ~\citet{loshchilov2016cma} compared the  Covariance Matrix Adaptation Evolution Strategy (CMA-ES) with Bayesian Optimization, and some of its variations, and achieved comparable results with Bayesian Optimization, and also with a lower computational cost.

In this paper, we propose a new population-based approach to searching in ANN's hyperparameter space. In particular, we use a population-based optimization algorithm, Biased Random Key Genetic Algorithm (BRKGA) to create a hybrid algorithm called HBRKGA by applying a Random Walk to each candidate solution. In our experiments, we compared HBRKGA to several other strategies: Grid Search, Random Search, Bayesian Optimization and CMA-ES. We used eight different datasets for classification. HBRKGA was able to increase the $F_1$ metric up to $1.2\%$ compared to the other methods, while showing a reasonable execution time.

This paper is organized as follows. Section~\ref{sec:background} defines the problem and shows basic concepts of optimization methods.
Section~\ref{sec:hbrkga} presents the proposed hybrid algorithm HBRKGA. 
Section~\ref{sec:experiments} indicates all experimental setups and the obtained results. Section~\ref{sec:conclusion} presents the conclusions.

\section{Background}
\label{sec:background}
We first present an overview to the problem of hyperparameter optimization in Section \ref{sec:problem-desc}. Then, we give a brief introduction to other optimization strategies: Grid Search (Section~\ref{sec:grid-search}), Random Search (Section~\ref{sec:random-search}),  Bayesian Optimization (Section~\ref{sec:bo}), and CMA-ES (Section~\ref{sec:cma}). After that,we introduce the main concepts related to Biased Random Key Genetic Algorithm (Section~\ref{sec:brkga}).

\subsection{Problem Statement}
\label{sec:problem-desc}

Let $\Gamma_\mathcal{A}$ be the set of possible hyperparameter combinations for a learning algorithm $\mathcal{A}$. Consider the general optimization problem of finding $\gamma^\star \in \Gamma_{\mathcal{A}}$, the best set of hyperparameters for $\mathcal{A}$. Here, \emph{best} is defined using a function $f: \Gamma_\mathcal{A} \rightarrow \Re$ that provides an estimate of the predictive performance of $\mathcal{A}$ on some validation set $\mathcal{X}$. This optimization problem can be formalized as:

\begin{equation}
\label{eq:optimization}
\gamma^\star = \argmin_{\gamma\in \Gamma_\mathcal{A}}f(\gamma; \mathcal{A}, \mathcal{X})
\end{equation}

The variable $\gamma$ is a vector sampled from hyperparameter space $\Gamma_\mathcal{A}$. The goal of the optimization problem is to find a $\gamma$ that minimizes $f$. Considering currently existing methods for hyperparameter search, there is no way to guarantee that the optimal value of $\gamma$ can be found. However, methods with a approximate approach (such as the ones in evidence in this work) can generate satisfactory models.

In this paper, we consider a particular family of learning algorithm, represented by feed-forward neural networks.

\subsection{Grid Search}
\label{sec:grid-search}

Grid Search performs a search by considering a multi-dimensional grid of hyperparameter combinations. The ranges for each hyperparameter in the grid are usually user-defined. Grid Search then computes a Cartesian product corresponding to the possible hyperparameter combinations \citep{bergstra2012random}. Since there may be some hyperparameters that can assume infinitely many values, the user must also define a step used to jump from one hyperparameter value to another. When done by a specialist in the domain in question, these sampled values can result in satisfactory learning models.

This technique has a trivial implementation and is easy to parallelize: each combination of hyperparameter values can be tested in parallel. However, the amount of hyperparameter combinations grows exponentially with the number of hyperparameters \citep{bellman1961adaptive,bergstra2012random}. As a result, Grid Search may exploit many unimportant areas in $\Gamma_{\mathcal{A}}$ if the input grid is not carefully designed by a domain expert. This problem causes a waste of computational resources since there is no rule to explore the hyperparameters space. 

\subsection{Random Search}
\label{sec:random-search}

Random Search takes as input a bounded subspace of $\Gamma_{\mathcal{A}}$. The bounds for such subspace are also user-provided. Then it takes random samples from this bounded domain, which are used as the hyperparameter combinations to be tested~\citep{zabinsky2009random}. It better suited to large-scale problems when compared to Grid Search, since avoid exploring less relevant areas in $\Gamma_{\mathcal{A}}$. There are some variants of this algorithm, such as using it in conjunction with the Two-Phased Method~\citep{schoen2002two}.

Compared to Grid Search, Random Search proves to be more efficient in the sense that it does not combine all selected values contained in a user-defined grid to perform hyperparameter optimization. Instead, it randomly explores regions of $\Gamma_{\mathcal{A}}$ that could have a better relevance in the hyperparameter space. This behavior makes it less computationally costly than Grid Search \citep{bergstra2012random}.

\subsection{Bayesian Optimization}
\label{sec:bo}

Both Random Search and Grid Search do not have a disciplined basis for optimizing hyperparameters. They perform search in a trial and error manner. Smarter hyperparameter optimization algorithms are able to generate a solution based on prior knowledge (i.e., previously generated solutions in the hyperparameter space). 
One of these methods is Bayesian Optimization. This method is able to make assumptions about the hyperparameter space based on prior samples to assist on choosing the next samples~\citep{brochu2010tutorial}. Bayesian optimization is able to construct a probabilistic model of a function to be optimized to explore the possible values of this function, using information obtained from previous iterations to generate a search model \citep{snoek2012practical}.

According to~\citet{snoek2012practical}, there are two components to be defined in Bayesian optimization. The first is the function that will express the assumption about the function to be optimized. Usually, the Gaussian Process is used to obtain samples through this function, but other models can be explored \citep{dewancker@baye}. The second component is the acquisition function, used to determine the next point to be explored. Some alternatives to this component are Probability of Improvement, Expected Improvement, and Upper Confidence Interval. 

Initially, several multivariate random samples are generated, where each of these is considered a different process. After that, Bayesian Optimization fits the function using the Gaussian Process. According to \citet{rasmussen2004gaussian}, a Gaussian Process is a generalization of Gaussian Probability Distribution. It generates a non-parametric model of probability to be able to estimate the unknown values of a function.

In Multivariate Random Distributions, it is necessary to use a covariance matrix to obtain the distance between points. This concept is the same as the standard deviation in a uni-variate normal distribution. There are several ways to calculate the distance between these points and to obtain this matrix. In the Gaussian process this function is called the kernel.

A Gaussian Process requires a specific kernel choice that best fits the data. The evaluation of these functions becomes computationally costly in a high-dimensional, multi-sampled data set. Some examples of kernels used in the Gaussian Process are: Linear, Exponential, Quadratic, Periodic, etc. These methods directly affect the distribution of data during the creation of the prior function.

\subsection{Covariance Matrix Adaptation Evolution Strategy}
\label{sec:cma}

Covariance Matrix Adaptation Evolution Strategy (CMA-ES) is a stochastic optimization method belonging to evolutionary algorithm family \citep{hansen2006cma}. CMA-ES is used to help in non-linear and non-convex problems \citep{loshchilov2014computationally}. Solutions are generated from normal multivariate distribution sampling, where the covariance matrix and mean of the sampling is adapted during each generation for the choice of candidates. The step size of the sampling is defined by the standard deviation, where its adaptation over generations can lead the algorithm to better convergence \citep{hansen2016cma}. CMA-ES proved to be a good alternative to Bayesian Optimization in the hyperparameter optimization problem. It is able to achieve comparable results and with less computational resources \citep{loshchilov2016cma}.

\subsection{Biased Random Key Genetic Algorithm}
\label{sec:brkga}

Genetic algorithms (GA) \citep{goldberg1988genetic} are methods that simulate the evolution of a population over a certain number of generations. Each individual in a population represents a candidate solution to a given optimization problem. These algorithms apply the concept of \emph{survival of the fittest individuals} to find reasonable quality solutions to optimization problems.
    
To escape from local optima and search in the solution space, a population evolves in several generations. The individuals or chromosomes represent a solution to the optimization problem. Each chromosome encodes a solution in a finite chain of bits or integers, enabling the definition of reproduction operators between two parents. An objective function expresses the fitness criterion in the selection of the right individuals.
    
A new population is generated from the combination of elements belonging to the current population at each generation of GA. It is performed by three principal operators: reproduction, crossover, and mutation. The new population is acquired as follows: i) The next population is sampled from a small percentage of the best individuals in the actual population; ii) crossover applies deterministic or probabilistic operators to randomly selected parents, generating offspring for the next generation; and iii) a random mutation of gene positions is performed to avoid local optima.
    
Random-key Genetic Algorithms (RKGA) were proposed by \citet{bean1994genetic}. In this method, vectors of decimal numbers whose values belong to the $[0,1]$ domain represent the chromosomes. Each vector is given as input to a deterministic algorithm called \textit{decoder}, which associates it with a solution of the optimization problem. The RKGA is an enhancement of the classic Genetic Algorithms, and its main objective is to mitigate GA operators' difficulty in dealing with feasible solutions. Thus, the representation of the problem parameters through random keys allows the development of problem independent operators.

\citet{gonccalves2011biased} developed Biased Random-Key Genetic Algorithms (BRKGA) based on RKGA. The main difference between BRKGA and RKGA is the biased way of how parents are chosen in the reproduction operator. To BRKGA obtain a new individual, the method combine an individual randomly chosen from an elite set ($p_e$) of the current population set ($p$) and another of a non-elite set $(p \setminus p_e)$ of individuals, in which $|p_e| < |p| - |p_e|$. A single individual can be selected more than one time and then can produce more than one offspring.

BRKGA needs $|p_e| < |p| - |p_e|$, the probability of an elite individual being selected for reproduction $(\frac{1}{|p_e|})$ is greater than that of a non-elite individual $(\frac{1}{|p|-|p_e|})$. So, elite individuals have greater probability of passing forward their characteristics to next generations. Moreover, the crossover concept in BRKGA is the Parameterized Uniform Crossing \citep{spears1995virtues} with $\Pr_e(i) > 0.5$, where $\Pr_e(i)$ is the probability that the $i$-th offspring inherits from an elite individual.

BRKGA has been successfully applied to several optimization problems such as Packaging \citep{gonccalves2007hybrid}, Routing \citep{martinez2011brkga}, Transmission Network Expansion Planning~\citep{gonzalez2018biased} and Traveling Salesman Problem Variants~\citep{samanlioglu2008memetic, snyder2006random}. In this work, we use the BRKGA meta-heuristic as a base for our proposed hybrid solution for hyperparameter optimization. In particular, our solution combines BRKGA with a Random-Walk procedure so that characteristics of the Random Search strategy could be harnessed.

\section{HBRKGA}
\label{sec:hbrkga}

In this section we describe HBRKGA, our proposed strategy for hyperparameter optimization. We begin by describing how a standard random key vector can be mapped into another vector corresponding to a hyperparameter configuration (Section~\ref{sec:hbrkga:enc:dec}). After that, we describe the Random-Walk procedure used by HBRKGA  (Section~\ref{sec:hbrkga:random:walk}). Finally, we present details related to the main procedure of our proposed strategy (Section~\ref{sec:hbrkga:main:proc}).

\subsection{Encoding and decoding candidate solutions}
\label{sec:hbrkga:enc:dec}

In general, each hyperparameter of a learning algorithm $\mathcal{A}$ has its corresponding range of values and data type. Furthermore, a given strategy for hyperparameter search may be able to search only a particular subrange of values of a hyperparameter. To cope with this, we define an abstract data type to be used in the procedures presented hereafter. Let us denote an instance of this abstract data type by $S_{\mathcal{A}}$. The data part of $S_{\mathcal{A}}$ holds a list. Each entry in this list corresponds to one hyperparameter of $\mathcal{A}$, in which its considered subrange and data type are stored. 

The operations $min(S_{\mathcal{A}}, i)$ and $max(S_{\mathcal{A}}, i)$ are defined for such an abstract data type. These operations return the minimum and maximum values for the $i$-th hyperparameter, respectively (i.e., its considered subrange). Another operation defined in the context of $S_{\mathcal{A}}$ is $dt(S_{\mathcal{A}}, i)$. this function returns the data type of the $i$-th hyperparameter. A final operation we define for this abstract data type is $round(S_{\mathcal{A}}, i, v)$. This operation returns the value that is closest to $v \in \Re$ considering two constraints: the returned value (1) has the same data type as the $i$-th hyperparameter and (2) it is inside the close interval $[min(S_{\mathcal{A}}, i), max(S_{\mathcal{A}}, i)]$.

Given the definition of the abstract data type provided above, we encode a solution in a vector $\bar{\gamma}$ of $n$ random keys, in which $n$ corresponds to the number of hyperparameters in $\mathcal{A}$. This way, a value for the $i$-th hyperparameter, $\gamma_i$ ($1 \leq i \leq n$), is mapped to its corresponding key $\bar{\gamma}_i$ using Eq.~\ref{eq:min-max}.

\begin{equation}
\label{eq:min-max}
\bar{\gamma}_i = \text{round}\left(S_{\mathcal{A}}, i, \frac{\gamma_i - min(S_{\mathcal{A}}, i)}{max(S_{\mathcal{A}}, i) - min(S_{\mathcal{A}}, i)}\right)
\end{equation}

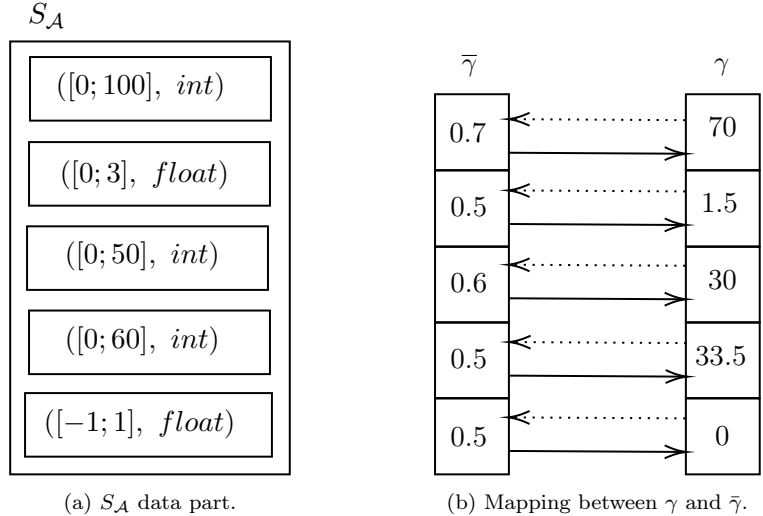
\begin{figure}[htp]
    \centering
    \begin{minipage}[b]{0.46\linewidth}
        \centering

\tikzset{every picture/.style={line width=0.75pt}} 

\begin{tikzpicture}[x=0.75pt,y=0.75pt,yscale=-1,xscale=1]

\draw   (90.5,69.75) -- (231.5,69.75) -- (231.5,288.25) -- (90.5,288.25) -- cycle ;
\draw   (100.5,77.61) -- (222,77.61) -- (222,109.71) -- (100.5,109.71) -- cycle ;
\draw   (100,120.55) -- (221.5,120.55) -- (221.5,152.64) -- (100,152.64) -- cycle ;
\draw   (99,163.06) -- (220.5,163.06) -- (220.5,195.15) -- (99,195.15) -- cycle ;
\draw   (100,205.57) -- (221.5,205.57) -- (221.5,237.66) -- (100,237.66) -- cycle ;
\draw   (98,247.23) -- (222.5,247.23) -- (222.5,279.32) -- (98,279.32) -- cycle ;

\draw (98,49) node [anchor=north west][inner sep=0.75pt]   [align=left] {$S_{\mathcal{A}}$};
\draw (112,83.92) node [anchor=north west][inner sep=0.75pt]   [align=left] {$\displaystyle ([ 0;100] ,\ int)$};
\draw (114,128.13) node [anchor=north west][inner sep=0.75pt]   [align=left] {$\displaystyle ([ 0;3] ,\ float)$};
\draw (116.5,168.94) node [anchor=north west][inner sep=0.75pt]   [align=left] {$\displaystyle ([ 0;50] ,\ int)$};
\draw (116.5,211.87) node [anchor=north west][inner sep=0.75pt]   [align=left] {$\displaystyle ([ 0;60] ,\ int)$};
\draw (106,253.88) node [anchor=north west][inner sep=0.75pt]   [align=left] {$\displaystyle ([ -1;1] ,\ float)$};

\end{tikzpicture}
\subcaption{$S_{\mathcal{A}}$ data part.}
\label{sa-data-part}
    \end{minipage}
    \begin{minipage}[b]{0.46\linewidth}
        \centering

\tikzset{every picture/.style={line width=0.75pt}} 

\begin{tikzpicture}[x=0.75pt,y=0.75pt,yscale=-1,xscale=1]

\draw   (52.25,69.89) -- (89.5,69.89) -- (89.5,108.16) -- (52.25,108.16) -- cycle ;

\draw   (52.25,108.42) -- (89.5,108.42) -- (89.5,146.69) -- (52.25,146.69) -- cycle ;

\draw   (52.25,146.94) -- (89.5,146.94) -- (89.5,185.21) -- (52.25,185.21) -- cycle ;

\draw   (52.25,184.96) -- (89.5,184.96) -- (89.5,223.22) -- (52.25,223.22) -- cycle ;

\draw   (52.25,223.48) -- (89.5,223.48) -- (89.5,261.75) -- (52.25,261.75) -- cycle ;

\draw   (178.75,69.89) -- (216,69.89) -- (216,108.16) -- (178.75,108.16) -- cycle ;

\draw   (178.75,108.42) -- (216,108.42) -- (216,146.69) -- (178.75,146.69) -- cycle ;

\draw   (178.75,146.94) -- (216,146.94) -- (216,185.21) -- (178.75,185.21) -- cycle ;

\draw   (178.75,184.96) -- (216,184.96) -- (216,223.22) -- (178.75,223.22) -- cycle ;

\draw   (178.75,223.48) -- (216,223.48) -- (216,261.75) -- (178.75,261.75) -- cycle ;

\draw [line width=0.75]    (89.5,99.42) -- (177,99.92) ;
\draw [shift={(179,99.93)}, rotate = 180.33] [color={rgb, 255:red, 0; green, 0; blue, 0 }  ][line width=0.75]    (10.93,-3.29) .. controls (6.95,-1.4) and (3.31,-0.3) .. (0,0) .. controls (3.31,0.3) and (6.95,1.4) .. (10.93,3.29)   ;
\draw [line width=0.75]  [dash pattern={on 0.84pt off 2.51pt}]  (178.5,82.98) -- (91,82.48) ;
\draw [shift={(89,82.46)}, rotate = 360.33000000000004] [color={rgb, 255:red, 0; green, 0; blue, 0 }  ][line width=0.75]    (10.93,-3.29) .. controls (6.95,-1.4) and (3.31,-0.3) .. (0,0) .. controls (3.31,0.3) and (6.95,1.4) .. (10.93,3.29)   ;
\draw [line width=0.75]    (89.5,135.39) -- (177,135.89) ;
\draw [shift={(179,135.9)}, rotate = 180.33] [color={rgb, 255:red, 0; green, 0; blue, 0 }  ][line width=0.75]    (10.93,-3.29) .. controls (6.95,-1.4) and (3.31,-0.3) .. (0,0) .. controls (3.31,0.3) and (6.95,1.4) .. (10.93,3.29)   ;
\draw [line width=0.75]  [dash pattern={on 0.84pt off 2.51pt}]  (178.5,118.95) -- (91,118.45) ;
\draw [shift={(89,118.43)}, rotate = 360.33000000000004] [color={rgb, 255:red, 0; green, 0; blue, 0 }  ][line width=0.75]    (10.93,-3.29) .. controls (6.95,-1.4) and (3.31,-0.3) .. (0,0) .. controls (3.31,0.3) and (6.95,1.4) .. (10.93,3.29)   ;
\draw [line width=0.75]    (89.5,172.88) -- (177,173.39) ;
\draw [shift={(179,173.4)}, rotate = 180.33] [color={rgb, 255:red, 0; green, 0; blue, 0 }  ][line width=0.75]    (10.93,-3.29) .. controls (6.95,-1.4) and (3.31,-0.3) .. (0,0) .. controls (3.31,0.3) and (6.95,1.4) .. (10.93,3.29)   ;
\draw [line width=0.75]  [dash pattern={on 0.84pt off 2.51pt}]  (178.5,156.45) -- (91,155.94) ;
\draw [shift={(89,155.93)}, rotate = 360.33000000000004] [color={rgb, 255:red, 0; green, 0; blue, 0 }  ][line width=0.75]    (10.93,-3.29) .. controls (6.95,-1.4) and (3.31,-0.3) .. (0,0) .. controls (3.31,0.3) and (6.95,1.4) .. (10.93,3.29)   ;
\draw [line width=0.75]    (90,211.92) -- (177.5,212.43) ;
\draw [shift={(179.5,212.44)}, rotate = 180.33] [color={rgb, 255:red, 0; green, 0; blue, 0 }  ][line width=0.75]    (10.93,-3.29) .. controls (6.95,-1.4) and (3.31,-0.3) .. (0,0) .. controls (3.31,0.3) and (6.95,1.4) .. (10.93,3.29)   ;
\draw [line width=0.75]  [dash pattern={on 0.84pt off 2.51pt}]  (179,195.49) -- (91.5,194.98) ;
\draw [shift={(89.5,194.97)}, rotate = 360.33000000000004] [color={rgb, 255:red, 0; green, 0; blue, 0 }  ][line width=0.75]    (10.93,-3.29) .. controls (6.95,-1.4) and (3.31,-0.3) .. (0,0) .. controls (3.31,0.3) and (6.95,1.4) .. (10.93,3.29)   ;
\draw [line width=0.75]    (89.5,249.94) -- (177,250.44) ;
\draw [shift={(179,250.45)}, rotate = 180.33] [color={rgb, 255:red, 0; green, 0; blue, 0 }  ][line width=0.75]    (10.93,-3.29) .. controls (6.95,-1.4) and (3.31,-0.3) .. (0,0) .. controls (3.31,0.3) and (6.95,1.4) .. (10.93,3.29)   ;
\draw [line width=0.75]  [dash pattern={on 0.84pt off 2.51pt}]  (178.5,233.5) -- (91,233) ;
\draw [shift={(89,232.98)}, rotate = 360.33000000000004] [color={rgb, 255:red, 0; green, 0; blue, 0 }  ][line width=0.75]    (10.93,-3.29) .. controls (6.95,-1.4) and (3.31,-0.3) .. (0,0) .. controls (3.31,0.3) and (6.95,1.4) .. (10.93,3.29)   ;

\draw (58.5,120.41) node [anchor=north west][inner sep=0.75pt]   [align=left] {0.5};
\draw (58.5,158.94) node [anchor=north west][inner sep=0.75pt]   [align=left] {0.6};
\draw (58.5,196.95) node [anchor=north west][inner sep=0.75pt]   [align=left] {0.5};
\draw (58.5,236.47) node [anchor=north west][inner sep=0.75pt]   [align=left] {0.5};
\draw (188.5,79.88) node [anchor=north west][inner sep=0.75pt]   [align=left] {70};
\draw (185,118.41) node [anchor=north west][inner sep=0.75pt]   [align=left] {1.5};
\draw (188.5,157.45) node [anchor=north west][inner sep=0.75pt]   [align=left] {30};
\draw (182,195.46) node [anchor=north west][inner sep=0.75pt]   [align=left] {33.5};
\draw (192,235.99) node [anchor=north west][inner sep=0.75pt]   [align=left] {0};
\draw (58,82.97) node [anchor=north west][inner sep=0.75pt]   [align=left] {0.7};
\draw (61.5,48.02) node [anchor=north west][inner sep=0.75pt]  [font=\small] [align=left] {\begin{minipage}[lt]{9.785608pt}\setlength\topsep{0pt}
\begin{center}
$\displaystyle \overline{\gamma }$
\end{center}

\end{minipage}};
\draw (189.5,50.59) node [anchor=north west][inner sep=0.75pt]  [font=\small] [align=left] {\begin{minipage}[lt]{8.765608pt}\setlength\topsep{0pt}
\begin{center}
$\displaystyle \gamma $
\end{center}

\end{minipage}};

\end{tikzpicture}

\subcaption{Mapping between $\gamma$ and $\bar{\gamma}$.}
\label{mapping-gamma-gammabar}
    \end{minipage}

\caption{Example of mapping (i.e., encoding or decoding) between a vector of hyperparameter values ($\gamma$) and a vector of BRKGA key values ($\bar{\gamma}$).}
    \label{fig:encoding}
\end{figure}

Figure~\ref{fig:encoding} illustrates the process of encoding and decoding candidate solutions through the transformation represented by Eq.~\ref{eq:min-max}. Figure~\ref{sa-data-part} shows the data part of an instance $S_{\mathcal{A}}$ of the abstract data type described in this section. This instance represents information about five hyperparameters. Figure~\ref{mapping-gamma-gammabar} shows the mapping between values of $\gamma$ (vector of hyperparameters) and $\bar{\gamma}$ (vector of random keys) according to the information in the instance $S_{\mathcal{A}}$ shown in Figure 1a.

\subsection{Random-Walk procedure}
\label{sec:hbrkga:random:walk}

A crucial component of HBRKGA is its Random-Walk procedure, whose pseudocode is presented in Algorithm~\ref{alg:random-walk}. Since it is impossible to analyze every solution in the neighborhood of $\bar{\gamma}$, in this work we chose to blindly explore the space in a small neighborhood. Hence, the Random-Walk phase performs a stochastic search in the neighborhood of a given individual $\bar{\gamma}$ of the current population, looking for a better candidate solution. In particular, a sequence of perturbations is generated from the original decoded solution and the best one is returned as the new best solution. Below, we provide details of such Random-Walk phase.

Algorithm~\ref{alg:random-walk} receives $\bar{\gamma}$ as an input parameter, a candidate solution for the optimization problem  containing $n$ random keys. In line \ref{alg:random-walk-decode}, the algorithm decodes the solution from BRKGA domain (random keys) to HBRKGA domain (hyperparameter values). The decoded solution is then passed to the $\operatorname{Evaluate}$ procedure (line~\ref{alg:randw-evaluate}) along with a learning algorithm $\mathcal{A}$ and a dataset $\mathcal{X}$. This procedure corresponds to apply $\mathcal{A}$ (using the values in $\gamma$ as hyperparameters) to fit a model to $\mathcal{X}$. This procedure returns a solution with its numeric value (i.e. score) that reflects the quality of the fitted model.

\begin{center}
\begin{algorithm}[htb]
    \Begin{
        Map $\bar{\gamma}$ to $\gamma$ using Eq.~\ref{eq:min-max};\label{alg:random-walk-decode}
        
    	$\gamma_{temp}.score \leftarrow \operatorname{Evaluate}(\gamma, \mathcal{A}, \mathcal{X})$\; \label{alg:randw-evaluate}
    	$\gamma_{temp} \leftarrow \gamma$\;
        \For{$1 \mbox{ \bf{to} }\operatorname{nmov}$\label{alg:randw-nmov}}{
            $\gamma_{temp} \leftarrow \operatorname{Movement}(\gamma_{temp}, S_{\mathcal{A}})$\;\label{alg:random-walk-movement}
            \label{alg:randw-movement}
            $\gamma_{temp}.score \leftarrow \operatorname{Evaluate}(\gamma_{temp},\mathcal{A}, \mathcal{X})$\;\label{alg:random-walk-each-step-eval}
            \If{$\gamma.score < \gamma_{temp}.score$\label{alg:random-walk-each-step-eval-if}}{
                $\gamma \leftarrow \gamma_{temp}$\;
                $\gamma.score \leftarrow \gamma_{temp}.score$\;
            }\label{alg:random-walk-each-step-eval-endif}
        }
        Map $\gamma$ to $\bar{\gamma}$ using Eq.~\ref{eq:min-max};\label{alg:randw-encode}
        
        \Return $\bar{\gamma}$\;
    }
    \caption{$\mbox{RandomWalk}(\bar{\gamma}, \operatorname{nmov},\mathcal{A}, \mathcal{X}, S_{\mathcal{A}})$}
    \label{alg:random-walk}
\end{algorithm}
\end{center}

The amount of steps to be performed in the random walk is determined by the input parameter $\operatorname{nmov}$ (line~\ref{alg:randw-nmov}). At each step a movement is performed (line~\ref{alg:random-walk-movement}) in the hyperparameter space. The definition of movement here corresponds to applying a perturbation to one of the components of the input vector. The component to which the perturbation is to be applied is chosen uniformly at random. The neighborhood considered in the exploration for the selected component $\gamma_i$ is the closed interval [$0, \gamma_i (1 + \epsilon)$], according to Eq.~\ref{eq:perturbation}. The small positive constant number $\epsilon$ is a hyperparameter of HBRKGA. We use the round function here (Section~\ref{sec:hbrkga:enc:dec}) to cope with the case in which the value resulting from the perturbation has an incompatible data type (e.g. a floating point value when $\gamma_i$ only assumes integer values).

\begin{equation}
\gamma_i \leftarrow \text{round}\left(S_{\mathcal{A}}, i, \gamma_i + (1-2 \times \text{Bernoulli}(0.5)) \times \text{Unif}(0, \gamma_i (1 + \epsilon)\right)
\label{eq:perturbation}    
\end{equation}

At each Random-Walk step, the resulting perturbed hyperparameters vector is evaluated (line~\ref{alg:random-walk-each-step-eval}) in order to keep track of the best current candidate solution (lines \ref{alg:random-walk-each-step-eval-if}-\ref{alg:random-walk-each-step-eval-endif}). At the end (line~\ref{alg:randw-encode}), the algorithm encodes the refined solution back to BRKGA domain (random keys) before returning it to HBRKGA main procedure.

\subsection{HBRKGA main procedure}
\label{sec:hbrkga:main:proc}

\begin{center}
\begin{algorithm}[H]
    \Begin{
        Initialize score of the best solution found: $\gamma^\star.score \gets \infty$\;
        
        Randomly generate a population $p$ with $q_{\text{ind}}$ $n$-dimensional vectors of random keys\; \label{alg:hbrkga-randomInit}
        
    	\While{stopping criterion not satisfied}{ \label{alg:hbrkga-init-p}
        	\For{$i \gets 1$ to $q_{\text{ind}}$}{ \label{alg:hbrkga-randomWalkLoop}
        	    $p[i] \gets \mbox{RandomWalk}(p[i],\operatorname{nmov},\mathcal{A}, \mathcal{X}, S_{\mathcal{A}})$
        	    \label{alg:hbrkga-solution}
        	}
            Partition $p$ into two sets: $p_e$ and $p_{\bar{e}}$\;\label{alg:hbrkga-partition}
            
            Initialize population of next generation: $p^+ \gets p_e$\;
            
            Generate set $p_m$ of mutants, each mutant with $n$ random keys\;\label{alg:hbrkga-mutants} 
            
            Add $p_m$ to population of next generation: $p^+ \gets p^+ \cup p_m$\;
            
            \For{$i \gets 1$ to $q_{\text{ind}}-(q_e+q_m)$}{ \label{alg:hbrkga-reproductionStart} 
                Select parent $a$ at random from $p_e$\;
                
                Select parent $b$ at random from $p_{\bar{e}}$\;
                
                \For{$j \gets 1$ to $n$} {\label{alg:hbrkga-drawReproduction} 
                    Draw random variable $X \sim \text{Bernoulli}(\phi_a)$

                    $c[j] \gets
                      \begin{cases}
                        a[j]     & \text{if X = 1}, \\
                        b[j] & \text{if X = 0}.
                      \end{cases}$

                }
                Add offspring $c$ to population of next generation: $p^+ \gets p^+ \cup \{c\}$\;            } \label{alg:hbrkga-reproductionEnd}
                
            Update population: $p \gets p^+$\; \label{alg:hbrkga-updatePopulation}
            
            Find best solution in $p$: $\gamma^+ \gets \argmin_{1 \leq i \leq q_{ind}} (p[i].score)$\; \label{alg:hbrkga-bestResultStart}
            
            \If{$\gamma^+.score < \gamma^\star.score$}{
                $\gamma^\star \gets \gamma^+$\;
                
            } \label{alg:hbrkga-BestResultEnd}
        }
        \Return $\gamma^\star$\;\label{alg:hbrkga-return}
    }
    \caption{$\operatorname{HBRKGA}(q_{\text{ind}},q_e,q_m,n,\phi_a,\operatorname{nmov},\mathcal{A}, \mathcal{X}, S_{\mathcal{A}})$}
    \label{alg:HBRKGA}
\end{algorithm}
\end{center}

Now that the Random-Walk phase has been described, we can proceed to present the main procedure of our proposed population-based approach to  hyperparameter search. Algorithm \ref{alg:HBRKGA} presents the pseudo-code for HBRKGA. The purpose of this algorithm is to find $\gamma^\star$, the best possible configuration of hyperparameter values for a given learning algorithm $\mathcal{A}$. 

A random initial population $p$ of individuals is generated (line~\ref{alg:hbrkga-randomInit}). Each individual $p[i]$ ($1 \leq i \leq q_{ind}$) in the population is a vector of random keys. Each component in such a vector is a random value drawn from a standard uniform distribution: $p[i,j] \sim \text{Unif}(0,1)$. 

The main loop of the algorithm starts at line~\ref{alg:hbrkga-init-p}. This loop is controlled by a stopping criterion. There are several alternative stopping criteria to use. Examples are the maximum number of generations, maximum runtime or until a specific value for the fitness function is reached. The inner loop starting at line~\ref{alg:hbrkga-randomWalkLoop} performs 
the Random-Walk phase (Algorithm~\ref{alg:random-walk}) for each individual. As a result, the $i$-th individual in $p$ is (potentially) changed to one of its neighbors, if it is the case that the latter evaluates better than the former.

Line~\ref{alg:hbrkga-partition} partitions the current population $p$ into two subsets $p_e$ and $p_{\bar{e}}$ in such a way that $|p_e| = q_e$ and $|p_{\bar{e}}| = q_{ind}-q_e$. To form $p_e$, the individuals in the current generation are first sorted according to their scores. Then, the top $q_e$ individuals from the current generation are selected to form the elite set $p_e$.

Line~\ref{alg:hbrkga-mutants} generates $q_m$ mutant individuals. The mutants replace a fraction of actual population with new random individuals with $n$ random keys. The mutants are created following a random uniform distribution. These individuals are important since they help the optimization process to escape local minima~\citep{gonccalves2011biased}.

The loop between lines~\ref{alg:hbrkga-reproductionStart}-\ref{alg:hbrkga-reproductionEnd} starts the reproduction operator. One random parent $a$ from $p_e$ and another one $b$ from $p_{\bar{e}}$ are selected. Then the inner loop in line~\ref{alg:hbrkga-drawReproduction} is executed for each key $n$. A random variable $X$ is generated to define which parent ($a$ or $b$) the offspring $c$ will inherit the characteristics of a specific key. The parent is selected according to a Bernoulli distribution with parameter $\phi_a$. This selection is biased towards the parent in the elite set (i.e., there is a greater probability of the parent in the elite set to be selected). After being created, the offspring $c$ is added to the next generation $p^+$.

After the reproduction process, the population $p$ is updated from $p^+$ in line~\ref{alg:hbrkga-updatePopulation}. Finally, find the best solution $y^+$ in the current population to update $\gamma^\star$ (only if $\gamma^+.score < \gamma^\star.score$) between lines~\ref{alg:hbrkga-bestResultStart}-\ref{alg:hbrkga-BestResultEnd}. The best solution ($\gamma^\star$) is finally returned (line~\ref{alg:hbrkga-return}).

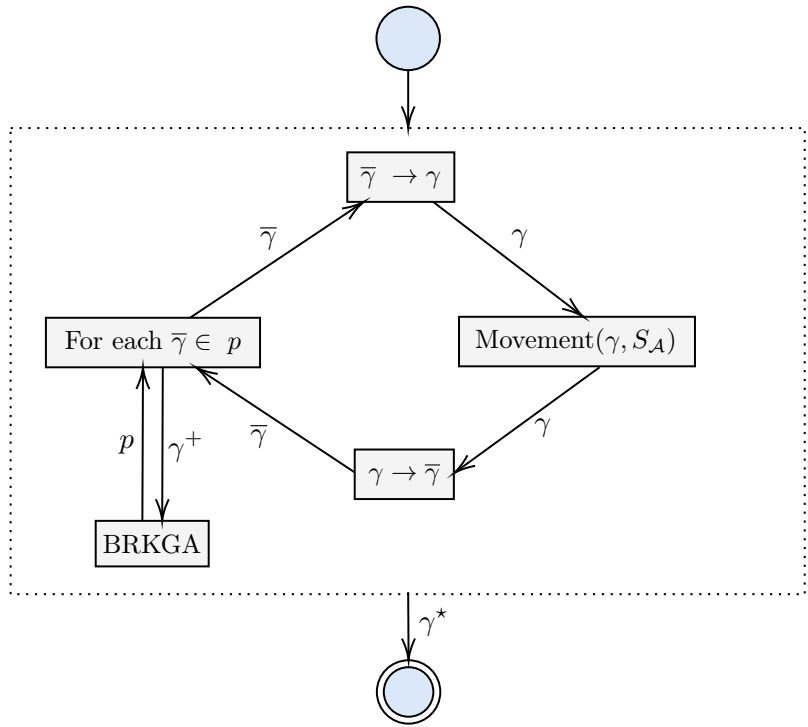
\begin{figure}[htb]
\centering

\tikzset{every picture/.style={line width=0.75pt}} 

\begin{tikzpicture}[x=0.75pt,y=0.75pt,yscale=-1,xscale=1]

\draw    (452,259.75) -- (380.12,312.07) ;
\draw [shift={(378.5,313.25)}, rotate = 323.95] [color={rgb, 255:red, 0; green, 0; blue, 0 }  ][line width=0.75]    (10.93,-3.29) .. controls (6.95,-1.4) and (3.31,-0.3) .. (0,0) .. controls (3.31,0.3) and (6.95,1.4) .. (10.93,3.29)   ;
\draw  [color={rgb, 255:red, 0; green, 0; blue, 0 }  ,draw opacity=1 ][fill={rgb, 255:red, 74; green, 144; blue, 226 }  ,fill opacity=0.2 ] (339.5,93.75) .. controls (339.5,84.91) and (346.66,77.75) .. (355.5,77.75) .. controls (364.34,77.75) and (371.5,84.91) .. (371.5,93.75) .. controls (371.5,102.59) and (364.34,109.75) .. (355.5,109.75) .. controls (346.66,109.75) and (339.5,102.59) .. (339.5,93.75) -- cycle ;
\draw  [color={rgb, 255:red, 0; green, 0; blue, 0 }  ,draw opacity=1 ][fill={rgb, 255:red, 255; green, 255; blue, 255 }  ,fill opacity=1 ] (339.7,423.6) .. controls (339.7,414.76) and (346.86,407.6) .. (355.7,407.6) .. controls (364.54,407.6) and (371.7,414.76) .. (371.7,423.6) .. controls (371.7,432.44) and (364.54,439.6) .. (355.7,439.6) .. controls (346.86,439.6) and (339.7,432.44) .. (339.7,423.6) -- cycle ;
\draw  [color={rgb, 255:red, 0; green, 0; blue, 0 }  ,draw opacity=1 ][fill={rgb, 255:red, 74; green, 144; blue, 226 }  ,fill opacity=0.2 ] (343.2,423.6) .. controls (343.2,416.7) and (348.8,411.1) .. (355.7,411.1) .. controls (362.6,411.1) and (368.2,416.7) .. (368.2,423.6) .. controls (368.2,430.5) and (362.6,436.1) .. (355.7,436.1) .. controls (348.8,436.1) and (343.2,430.5) .. (343.2,423.6) -- cycle ;

\draw    (355.5,373.25) -- (355.69,405.6) ;
\draw [shift={(355.7,407.6)}, rotate = 269.67] [color={rgb, 255:red, 0; green, 0; blue, 0 }  ][line width=0.75]    (10.93,-3.29) .. controls (6.95,-1.4) and (3.31,-0.3) .. (0,0) .. controls (3.31,0.3) and (6.95,1.4) .. (10.93,3.29)   ;
\draw  [dash pattern={on 0.84pt off 2.51pt}] (155,139) -- (555.5,139) -- (555.5,374.25) -- (155,374.25) -- cycle ;
\draw    (355.5,109.75) -- (355.5,137) ;
\draw [shift={(355.5,139)}, rotate = 270] [color={rgb, 255:red, 0; green, 0; blue, 0 }  ][line width=0.75]    (10.93,-3.29) .. controls (6.95,-1.4) and (3.31,-0.3) .. (0,0) .. controls (3.31,0.3) and (6.95,1.4) .. (10.93,3.29)   ;
\draw    (328.5,312.75) -- (250.17,260.85) ;
\draw [shift={(248.5,259.75)}, rotate = 393.52] [color={rgb, 255:red, 0; green, 0; blue, 0 }  ][line width=0.75]    (10.93,-3.29) .. controls (6.95,-1.4) and (3.31,-0.3) .. (0,0) .. controls (3.31,0.3) and (6.95,1.4) .. (10.93,3.29)   ;

\draw  [color={rgb, 255:red, 0; green, 0; blue, 0 }  ,draw opacity=1 ][fill={rgb, 255:red, 202; green, 200; blue, 200 }  ,fill opacity=0.2 ]  (197.9,337.24) -- (254.9,337.24) -- (254.9,360.24) -- (197.9,360.24) -- cycle  ;
\draw (226.4,348.74) node  [font=\fontsize{0.93em}{1.12em}\selectfont] [align=left] {BRKGA};
\draw  [color={rgb, 255:red, 0; green, 0; blue, 0 }  ,draw opacity=1 ][fill={rgb, 255:red, 202; green, 200; blue, 200 }  ,fill opacity=0.2 ]  (172.79,234.74) -- (280.79,234.74) -- (280.79,259.74) -- (172.79,259.74) -- cycle  ;
\draw (226.79,247.24) node  [font=\fontsize{0.93em}{1.12em}\selectfont] [align=left] {For each $\displaystyle \overline{\gamma } \in \ p$};
\draw  [color={rgb, 255:red, 0; green, 0; blue, 0 }  ,draw opacity=1 ][fill={rgb, 255:red, 202; green, 200; blue, 200 }  ,fill opacity=0.2 ]  (324.78,151.24) -- (378.78,151.24) -- (378.78,176.24) -- (324.78,176.24) -- cycle  ;
\draw (351.78,163.74) node  [font=\fontsize{0.93em}{1.12em}\selectfont] [align=left] {$\displaystyle \overline{\gamma } \ \rightarrow \gamma $};
\draw  [color={rgb, 255:red, 0; green, 0; blue, 0 }  ,draw opacity=1 ][fill={rgb, 255:red, 202; green, 200; blue, 200 }  ,fill opacity=0.2 ]  (381.14,234.24) -- (500.14,234.24) -- (500.14,259.24) -- (381.14,259.24) -- cycle  ;
\draw (440.14,246.74) node  [font=\fontsize{0.93em}{1.12em}\selectfont] [align=left] {Movement$(\gamma, S_{\mathcal{A}})$};
\draw  [color={rgb, 255:red, 0; green, 0; blue, 0 }  ,draw opacity=1 ][fill={rgb, 255:red, 202; green, 200; blue, 200 }  ,fill opacity=0.2 ]  (328.53,301.24) -- (378.53,301.24) -- (378.53,326.24) -- (328.53,326.24) -- cycle  ;
\draw (353.53,313.74) node  [font=\fontsize{0.93em}{1.12em}\selectfont] [align=left] {$\displaystyle \gamma \rightarrow \overline{\gamma }$};
\draw (208.47,292.6) node [anchor=north west][inner sep=0.75pt]   [align=left] {$\displaystyle p$};
\draw (279.33,187) node [anchor=north west][inner sep=0.75pt]   [align=left] {$\displaystyle \overline{\gamma }$};
\draw (406.17,188.5) node [anchor=north west][inner sep=0.75pt]   [align=left] {$\displaystyle \gamma $};
\draw (417.75,284) node [anchor=north west][inner sep=0.75pt]   [align=left] {$\displaystyle \gamma $};
\draw (274.5,286.83) node [anchor=north west][inner sep=0.75pt]   [align=left] {$\displaystyle \overline{\gamma }$};
\draw (233,290.33) node [anchor=north west][inner sep=0.75pt]   [align=left] {$\displaystyle \gamma ^{+}$};
\draw (359.7,111.27) node [anchor=north west][inner sep=0.75pt]   [align=left] {};
\draw (359.5,379.75) node [anchor=north west][inner sep=0.75pt]   [align=left] {$\displaystyle \gamma ^{\star }$};
\draw    (245.5,234.74) -- (331.41,177.35) ;
\draw [shift={(333.07,176.24)}, rotate = 506.26] [color={rgb, 255:red, 0; green, 0; blue, 0 }  ][line width=0.75]    (10.93,-3.29) .. controls (6.95,-1.4) and (3.31,-0.3) .. (0,0) .. controls (3.31,0.3) and (6.95,1.4) .. (10.93,3.29)   ;
\draw    (368.1,176.24) -- (442.24,233.03) ;
\draw [shift={(443.82,234.24)}, rotate = 217.45] [color={rgb, 255:red, 0; green, 0; blue, 0 }  ][line width=0.75]    (10.93,-3.29) .. controls (6.95,-1.4) and (3.31,-0.3) .. (0,0) .. controls (3.31,0.3) and (6.95,1.4) .. (10.93,3.29)   ;
\draw    (221.44,337.24) -- (221.74,261.74) ;
\draw [shift={(221.74,259.74)}, rotate = 450.22] [color={rgb, 255:red, 0; green, 0; blue, 0 }  ][line width=0.75]    (10.93,-3.29) .. controls (6.95,-1.4) and (3.31,-0.3) .. (0,0) .. controls (3.31,0.3) and (6.95,1.4) .. (10.93,3.29)   ;
\draw    (231.74,259.74) -- (231.45,335.24) ;
\draw [shift={(231.44,337.24)}, rotate = 270.22] [color={rgb, 255:red, 0; green, 0; blue, 0 }  ][line width=0.75]    (10.93,-3.29) .. controls (6.95,-1.4) and (3.31,-0.3) .. (0,0) .. controls (3.31,0.3) and (6.95,1.4) .. (10.93,3.29)   ;

\end{tikzpicture}

\caption{HBRKGA overview. Individuals of the current generation $p$ are created according to BRKGA rules. Then each individual is possibly refined in the Random-Walk procedure.}
\label{fig:hbrkga-flux}
\end{figure}

Figure \ref{fig:hbrkga-flux} presents a birds-eye view of HBRKGA. Initially, HBRKGA receives a hyperparameter space $\S_\mathcal{A}$ and sends an individual solution to decode the individual $\bar{\gamma}$ to a value $\gamma$ mapped to the domain in question. Using a Random Walk, a local search like method is applied $\text{nmov}$ times to the solution $\gamma$. After that, $\gamma$ is encoded and the best individual in the current population $\gamma ^{+}$ is returned to HBRKGA framework, as presented in Algorithm \ref{alg:HBRKGA}. This process is repeated for each generation, until the return of the final best solution $\gamma^{\star}$.

\section{Experiments}
\label{sec:experiments}

In this section we describe the computational experiments we performed to validate our proposed hyperparameter optimization method. We perform experiments on eight publicly available datasets coming from several different application domains. We start by providing details about the datasets (Section~\ref{sec:datasets}), evaluation metrics (Section~\ref{sec:EvaluationMetrics}), and experimental settings (Section~\ref{sec:HyperparameterDistribution}). Further, we describe the main results of the experiments for each dataset (Section~\ref{sec:exp-results}) and summarize the results of ablation studies (Section~\ref{sec:ablation}).

\subsection{Datasets}
\label{sec:datasets}

To perform our validation experiments, we used eight datasets in total. Six of them are provided by \citet{larochelle2007empirical}. These are the original version and variants of MNIST, one of the most popular datasets in the image recognition and classification areas. We also reuse the Rectangles dataset from~\citet{larochelle2007empirical}. We also used the Fashion-MNIST dataset~\citep{xiao2017fashion}. Finally, to provide a better basis for experimental results, we have added COSMOS, an unbalanced dataset~\citep{fadely2012star, machado2016exploring}. 

\subsubsection{MNIST}
\label{subsec:mnist}

MNIST is a set of handwritten digit image data, having 60,000 examples in the training set and 10,000 examples in the test set \citep{lecun1998mnist}. The images in the dataset have a size of 28x28 pixel, totalizing 784 features. Several studies have already done using this dataset \citep{bergstra2012random, larochelle2009exploring}, one of the objectives of these studies being to achieve the smallest possible error in the identification of these digits.

\begin{figure}[htb]
    \centering
    \includegraphics[width=0.95\linewidth]{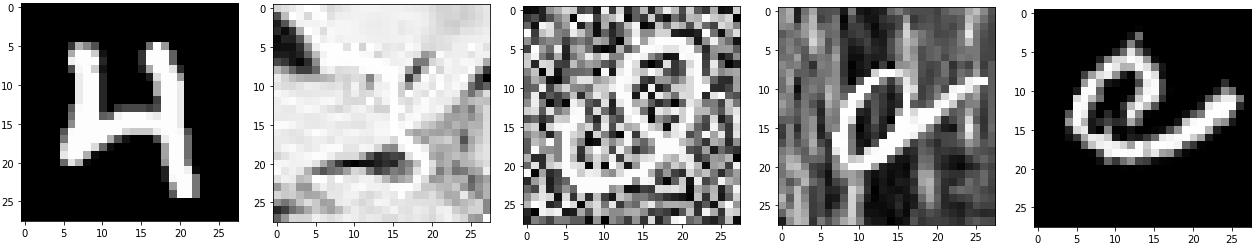}
    \caption{MNIST and its variations.}
    \label{fig:MNIST_variations}
\end{figure}

\citet{larochelle2007empirical} present a study with many factors of variations on top of MNIST, such as rotated digits and the addition of noise in the background of the images. With these variations, it is possible to observe several factors in the classification. These datasets are also used by \citet{bergstra2012random} to perform Random Search experiments. The MNIST variations that we selected are shown below and can be observed some examples in Figure~\ref{fig:MNIST_variations}. We follow the split: 12,000 images for training (the last 2,000 examples was used in validation set) and 50,000 images for test.

\begin{enumerate}
\item MNIST rotated (MNIST-R): the images suffered slight rotation in the digits, trying to reproduce different writing styles.
\item MNIST with random background (MNIST-RanBack): adding a random background in the digit images. This factor produces noises in the digits.
\item MNIST with image background (MNIST-IB): a background was produced with pieces of 20 images taken from the Internet.
\item MNIST with rotation and background (MNIST-RotBack): combination of the first two MNIST variations that were presented, resulting in rotate digits with some noise in the background.
\end{enumerate}

\subsubsection{Rectangles}
\label{subsec:rectangles}

\begin{figure}[htb]
    \centering
    \includegraphics[width=0.6\linewidth]{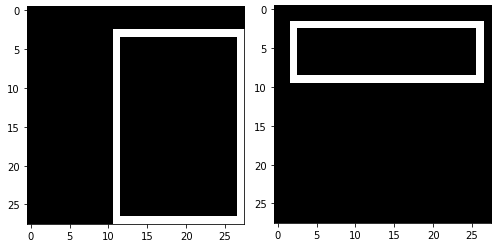}
    \caption{Tall and wide rectangle example.}
    \label{fig:rectangles_variations}
\end{figure}

In addition to MNIST and its variations, we selected one more case from~\citet{larochelle2007empirical}, the rectangles images. The objective of the rectangles dataset is the discrimination between tall and wide rectangles. Like MNIST, it has 28x28 pixel dimensions. The Figure~\ref{fig:rectangles_variations} shows an example of the label tall and wide. The training set has 1000 images and the validation set has 200 images. The test set has 50000 images.

\subsubsection{Fashion-MNIST}
\label{subsec:fashion-mnist}

\begin{figure}[htb]
    \centering
    \includegraphics[width=0.90\linewidth]{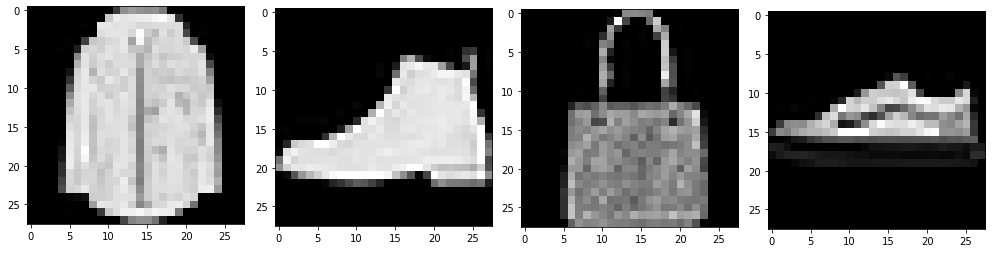}
    \caption{Fashion-MNIST images example.}
    \label{fig:fashion-mnist}
\end{figure}

The Fashion-MNIST dataset provided by \citet{xiao2017fashion} contains 60,000 training examples and 10,000 test examples in 28x28 grayscale images divided by 10 categories of fashion products: t-shirt, trouser, pullover, dress, coat, sandals, shirt, sneaker, bag and ankle boots. These classes are balanced over this dataset. It is an alternative to the MNIST benchmark for machine learning algorithms with more complex tasks for the correct classification. For the validation set, we use 10\% of the training set. Figure~\ref{fig:fashion-mnist} shows some examples of images from this dataset.

\subsubsection{COSMOS dataset}
\label{subsec:cosmos}

The COSMOS (Cosmic Evolution Survey) dataset \citep{scoville2007cosmic} is a catalog with information about more than 500000 astronomical objects and 90 attributes with their photometric measures. We used the same dataset and preprocessing as in \citet{machado2016exploring}, that takes into account feature selection, outliers removal, data cleaning, normalization and test/validation split. It is used in the problem knows as star/galaxy separation. This is a highly unbalanced dataset (386,957 stars and 5,542  galaxies after preprocessing, 98.55\% of them are stars). We used five photometric  features and its related error in the measure as the input for the ANN, totalling 10 features and a target label for star/galaxy classification.

\subsection{Evaluation Metric}
\label{sec:EvaluationMetrics}

In the experiments presented in this paper, we restrict ourselves to classification problems. Hence, we selected the metric $F_1$-score to measure models' quality in all the evaluated hyperparameter search methods. The $F_1$ metric (Eq.~\ref{eq:prec_rec}) is computed as the harmonic mean of two other metrics, precision and recall~\citep{han2011data}. Precision is the percentage of examples predicted by the model as belonging to a given class that genuinely belong to that class (Eq.~\ref{eq:precision}).  Recall is the percentage of examples of a given class correctly classified as so by the model (Eq.~\ref{eq:recall}). 

\noindent\begin{minipage}{.32\linewidth}
  \begin{equation}
    \label{eq:precision}
    \pi=\frac{\text{TP}}{\text{TP} + \text{FP}}
  \end{equation}
\end{minipage}
\begin{minipage}{.32\linewidth}
  \begin{equation}
    \label{eq:recall}
    \rho=\frac{\text{TP}}{\text{TP} + \text{FN}}
  \end{equation}
\end{minipage}
\begin{minipage}{.3\linewidth}
  \begin{equation}
    \label{eq:prec_rec}
    F_1 = \frac{2 \pi \rho}{\pi + \rho}
  \end{equation}
\end{minipage}

\vspace{0.5cm}

In the equations above, TP, FP, and FN are the true positive, false positive, and false negative counts. The $F_1$ score ranges from 0 to 1. A good model is expected to achieve an $F_1$ value close to 1, while models with low predictive quality tend to produce an $F_1$ score near 0. Since all the datasets we use in our experiments present multiple classes, we simply average the $F_1$-scores for each class and calculate a mean $F_1$-score as the final evaluation metric.

\subsection{Experimental Settings}
\label{sec:HyperparameterDistribution}

We ran the experiments on a computer with an Intel(R) Core(TM) i7-6700 CPU 3.40GHz processor, 32GB RAM, equipped with a GeForce GTX 1080 GPU. The ANN algorithm was developed using the \textit{Tensorflow} library~\citep{abadi2016tensorflow}. As a basis to develop HBRKGA, we used a BRKGA implementation provided by~\citet{toso2015c++}. We implemented the abstract data type described in Section~\ref{sec:hbrkga:enc:dec} to cope with the problem-dependent mapping procedure for converting a vector of random keys into hyperparameter values and vice-versa.

Inspired by previous similar experimental work, namely \citet{bergstra2012random} and \citet{larochelle2007empirical}, we selected five hyperparameters in a range of values to be explored by the methods presented. They are based on an ANN with three hidden layers architecture. The reused hyperparameters and their values are presented in three initial lines in Table~\ref{tab:hyper-space}. In particular, we chose the following hyperparameters: number of neurons in first, second, and third hidden layers, learning rate, and regularization rate. For the COSMOS dataset, we defined the range proportionally, since this dataset has a much lower number of input features.

\begin{table}[htb]
\centering
\caption{Hyperparameter range values for each dataset and its variants.}
\label{tab:hyper-space}
\begin{adjustbox}{width=\textwidth}
\begin{tabular}{lccccc@{\extracolsep{\fill}}} 
\toprule
\cmidrule{5-6}                            
Dataset         & \parbox{1.5cm}{\centering Neurons Layer 1} 
   & \parbox{1.5cm}{\centering Neurons Layer 2}   & \parbox{1.5cm}{\centering Neurons Layer 3}  & \parbox{1.5cm}{\centering Learning rate}  & \parbox{1.5cm}{\centering Reg} 
 \\ 
\midrule
\multirow{1}{*}{\parbox{3cm}{MNIST \\ }} 
    & [1000, 2000]       & [2000, 4000]  & [2000, 6000] & [$10^{-6}$, $10^{-1}$]           & [0, $10^{-3}$]      \\

\midrule

\multirow{1}{*}{\parbox{3cm}{Fashion-MNIST \\ }} 
    & [1000, 2000]       & [2000, 4000]  & [2000, 6000] & [$10^{-6}$, $10^{-1}$]           & [0, $10^{-3}$]      \\

\midrule
\multirow{1}{*}{\parbox{3cm}{Rectangles \\ }} 
    & [1000, 2000]       & [2000, 4000]  & [2000, 6000] & [$10^{-6}$, $10^{-1}$]& [0, $10^{-3}$] \\
\midrule
\multirow{1}{*}{\parbox{3cm}{COSMOS \\ }} 
    & [5, 15]       & [5, 30]  & [5, 45] & [$10^{-6}$, $10^{-1}$]& [0, $10^{-3}$] \\
\bottomrule
\end{tabular}
\end{adjustbox}
\end{table}

We use the optimization strategies described in Section~\ref{sec:background} as baseline for comparison to HBRKGA. We implemented Grid Search and Random Search from scratch. We used publicly available implementations for Bayesian Optimization\footnote{\url{https://github.com/fmfn/BayesianOptimization}} and CMA-ES\footnote{\url{https://github.com/CMA-ES/pycma}}.

We keep track of the number of solutions produced by each optimization strategy in each run of experiments for time comparison between them. The Grid Search optimization generates 240 combinations of different hyperparameters values to run in each dataset. This number results from combining the following values: 2 values for the first layer, three values for the second layer, four values for the third layer, five values for the learning rate, and two values for regularization rate (denoted Reg in Table~\ref{tab:hyper-space}). Due to that, and to make fair comparisons, we limited the maximum number of searches (i.e., a generation of hyperparameters) in each strategy to 240, already including the initial solution performed by Bayesian Optimization, CMA-ES, and HBRKGA. In this work, we configure HBRKGA parameters according to Table~\ref{tab:hbrkga-parameters}. In Bayesian Optimization, we use Upper Confidence Bound as acquisition function with 20 random initial points and 220 optimization steps. Finally, at CMA-ES we use 10 generations with 24 individuals.

\begin{table}[ht]
\centering
\begin{tabular}{lc}
\toprule
 \multicolumn{1}{c}{Parameter} &
 \begin{tabular}{c}Value \end{tabular} \\
\midrule
    Max. number of populations (stopping criteria) &   10\\
    Population size ($q_{ind}$) &   6\\
    Elite set size  ($q_{e}$) &   $2$\\
    Mutant set size ($q_{m}$) &   $1$\\
    Offspring probability ($\phi_a$)  &   0.7\\
    Steps in Random-Walk (nmov)  &   3\\
    Perturbation ratio ($\epsilon$) &  $15\%$ \\

\bottomrule
\end{tabular}

\caption{HBRKGA parameters settings.}
\label{tab:hbrkga-parameters}
\end{table}

We use cross-entropy as loss function, with a softmax activation function as output layer. In each hidden layer, the ReLU activation function is used. To save computational resources, we used early stopping technique. The goal is to stop the network training process when the value for the loss function does not decrease for a number of consecutive epochs. We configured the training process to generate a maximum of 300 epochs. If in 13 consecutive epochs a certain the loss function does no decreased in the validation set, the training is automatically stopped, the best model found is returned. We also use ADAM \citep{kingma2014adam} optimizer for training the ANNs.

\subsection{Experimental Results}
\label{sec:exp-results}

For each dataset, we performed ten runs of experiments for each hyperparameter optimization strategy covered in this work. We then computed statistical summaries for the $F_1$ metric and the execution time (in seconds). We divide the presentation of the experimental results into two parts. In the first part, we describe results related to the predictive quality of the classification models produced using each search strategy (Section~\ref{sec:exp-results:pred:quali}). In the second part, we present computational performance results concerning each strategy  (Section~\ref{sec:exp-results:comp:perf}).

\subsubsection{Predictive Quality}
\label{sec:exp-results:pred:quali}

Table~\ref{tab:f1-results} presents the results obtained by taking the mean and standard deviation of the best $F_1$ value found in each of the ten trials of experiments in the validation set for each dataset. We observe an increase in the mean of HBRKGA compared to Bayesian Optimization, CMA-ES, Random Search, and Grid Search in 6 of 8 datasets. The CMA-ES method was able to outperform HBRKGA results on MNIST-IB and matched the HRBKGA results on MNIST-RotBack. The most significant difference in the mean of $F_1$ between the HBRKGA and the second-best method occurred in the COSMOS and MNIST-R datasets. In these cases, HBRKGA increased the mean $F_1$ value by 0.006 and 0.009, respectively. Only in MNIST-RandBack, the Bayesian Optimization method was able to overcome the CMA-ES.

To summarize, HBRKGA obtained the best average $F_1$, followed by CMA-ES and Bayesian Optimization. In this global metric, Grid Search and Random Search achieved the worst $F_1$ averages among the methods tested. It is also possible to observe that the CMA-ES presented the lowest global mean in the standard deviation value.

\begin{table}[htb]
\centering
\caption{Average $F_1$ results for 10 experimental runs for each dataset. The best results are presented in bold face. The last line presents the average results considering all ten runs.}
\label{tab:f1-results}
\resizebox{\columnwidth}{!}{%
\begin{tabular}{@{}ccccccccccccccc@{}}
\toprule
\multicolumn{1}{l}{} 
& \multicolumn{2}{c}{GS} 
& \multicolumn{2}{c}{RS}
& \multicolumn{2}{c}{BO}
& \multicolumn{2}{c}{CMA-ES}
& \multicolumn{2}{c}{HBRKGA} \\

\cmidrule(lr){2-3} \cmidrule(lr){4-5} \cmidrule(lr){6-7} \cmidrule(lr){8-9} \cmidrule(lr){10-11}
 
& avg
& std
& avg
& std
& avg
& std
& avg
& std
& avg
& std
\\ \midrule

\multirow{1}{*}{MNIST} & 0.958  & 0.0047 & 0.962 & 0.0025 & 0.960 
                       & 0.0029 & 0.962 & 0.0009   & \textbf{0.965}    & 0.0014   \\ [0.5ex]
\multirow{1}{*}{MNIST-R}  & 0.877 & 0.0020 & 0.877 & 0.0175 & 0.879
                          & 0.0037 & 0.882 & 0.0021 & \textbf{0.891} & 0.0017\\[0.5ex]
\multirow{1}{*}{MNIST-IB} & 0.729 & 0.0121 & 0.741 & 0.0104 & 0.742
                          & 0.0158 & \textbf{0.748} & 0.0093 & 0.746 & 0.0082\\[0.5ex]
\multirow{1}{*}{MNIST-RotBack} & 0.358 & 0.0051 & 0.345 & 0.0049 & 0.359
                               & 0.0046 & \textbf{0.365} & 0.0050 & \textbf{0.365} & 0.0039\\[0.5ex]
\multirow{1}{*}{MNIST-RandBack} & 0.727 & 0.0033 & 0.708 & 0.0031 & 0.735
                                & 0.0391 & 0.731 & 0.0101 & \textbf{0.73}6 & 0.0154\\[0.5ex]
\multirow{1}{*}{Fashion-MNIST}   & 0.859 & 0.0007 & 0.860 & 0.0010 & 0.865 & 0.0037 & 0.865 & 0.0019 & \textbf{0.867} & 0.0035\\[0.5ex]
\multirow{1}{*}{Rectangles} & 0.965 & 0.0051 & 0.972 & 0.0045 & 0.975
                            & 0.0034 & 0.977 & 0.0036 & \textbf{0.981} & 0.0031\\[0.5ex]
\multirow{1}{*}{COSMOS}   & 0.757 & 0.0161 & 0.761 & 0.0126 & 0.761
                          & 0.0049 & 0.771 & 0.0107 & \textbf{0.777} & 0.0129\\[0.5ex]\hline
 \multirow{1}{*}{}  & 0.778 & 0.0061 & 0.778 &  0.0070 & 0.784 & 0.0097 & 0.787 & 0.0054 & \textbf{0.791} & 0.0062                    \\[0.5ex]
\end{tabular}%
}
\end{table}

\begin{figure}[htp]
    \centering
    \begin{minipage}[b]{0.46\linewidth}
        \centering
        \includegraphics[width=\textwidth]{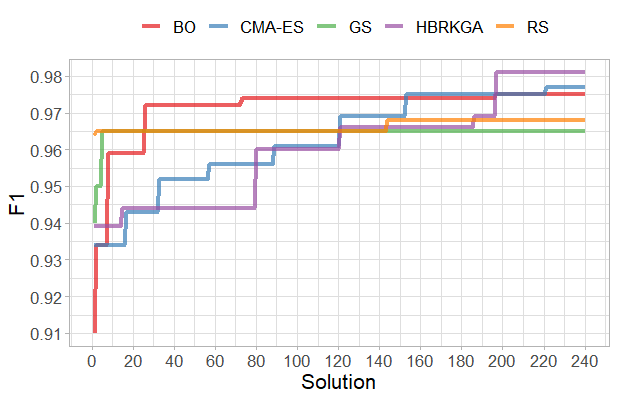}
        \subcaption{Rectangles}
    \end{minipage}
    \begin{minipage}[b]{0.46\linewidth}
        \centering
        \includegraphics[width=\textwidth]{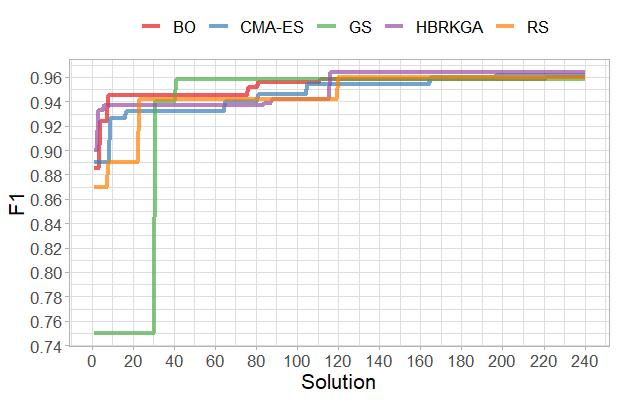}
        \subcaption{MNIST}
    \end{minipage}
        \begin{minipage}[b]{0.46\linewidth}
        \centering
        \includegraphics[width=\textwidth]{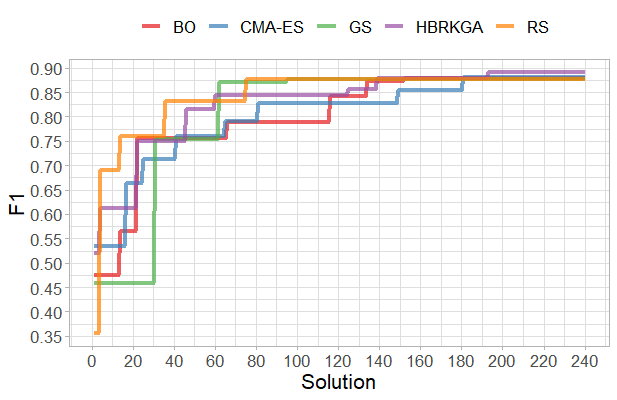}
        \subcaption{MNIST-R}
    \end{minipage}
        \begin{minipage}[b]{0.46\linewidth}
        \centering
        \includegraphics[width=\textwidth]{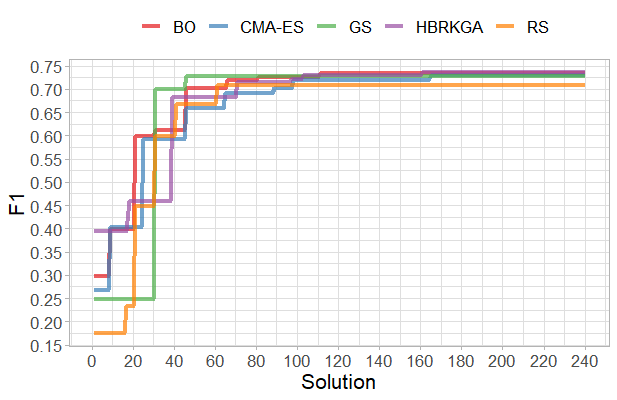}
        \subcaption{MNIST-RandBack}
    \end{minipage}
        \begin{minipage}[b]{0.4546\linewidth}
        \centering
        \includegraphics[width=\textwidth]{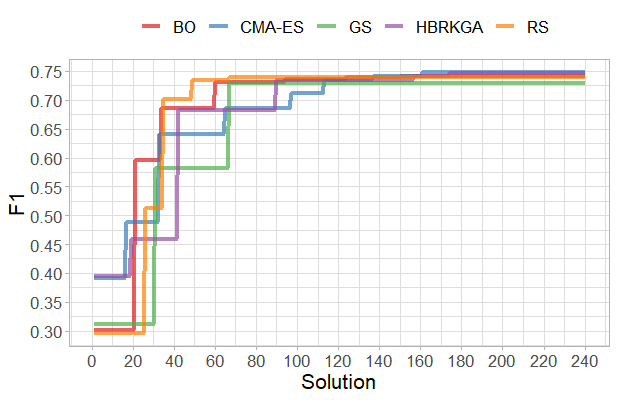}
        \subcaption{MNIST-IB}
    \end{minipage}
        \begin{minipage}[b]{0.46\linewidth}
        \centering
        \includegraphics[width=\textwidth]{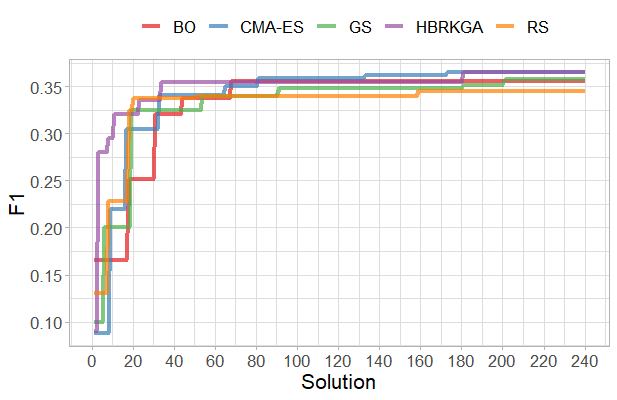}
        \subcaption{MNIST-RotBack}
    \end{minipage}
        \begin{minipage}[b]{0.46\linewidth}
        \centering
        \includegraphics[width=\textwidth]{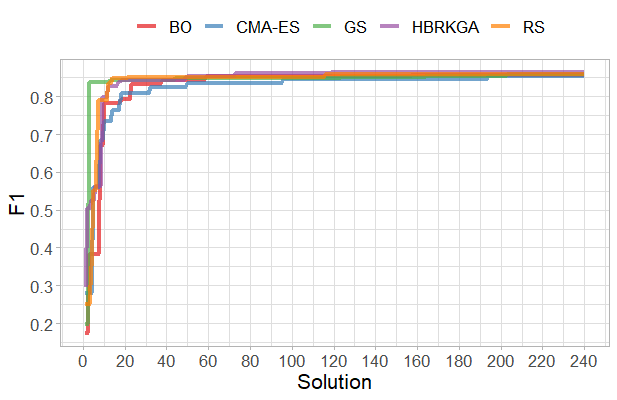}
        \subcaption{Fashion-MNIST}
    \end{minipage}
        \begin{minipage}[b]{0.46\linewidth}
        \centering
        \includegraphics[width=\textwidth]{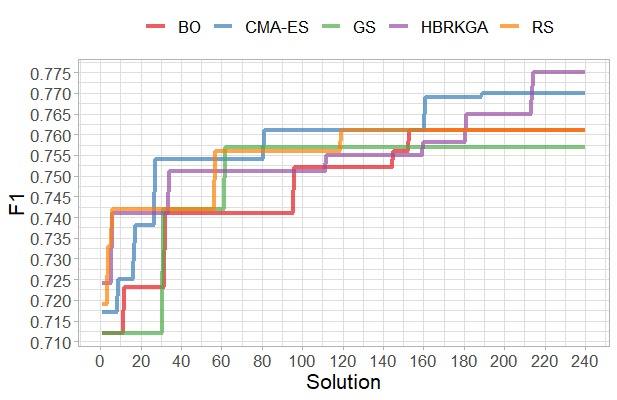}
        \subcaption{Cosmos}
    \end{minipage}

    \caption{$F_1$ mean evolution curve of each method for each dataset.}
    \label{fig:evolutionplot}
\end{figure}

Figure \ref{fig:evolutionplot} shows the average $F_1$ value of the solutions found by the algorithms over time. The evolution of both Grid Search and Random Search (green and orange lines) is quick, since they reach they corresponding maxima earlier than the other strategies. After this, they remain practically constant over time. However, in most cases the maximum found by these strategies if lower than the ones found in other strategies. A possible reason for this is that those two strategies do not have a way to escape local mimima. On the other hand, Bayesian Optimization, CMA-ES and HBRKGA (red, blue and purple lines) present, in most of the datasets studied, a constant evolution in the value of $F_1$ along the generated solutions. This common characteristic of these methods was determinant factors for the overall better results they presented. Also, it is possible to notice that Grid Search and Random Search strategies, despite not being able to produce the best solutions, need approximately 100 solutions to find their best $F_1$ average result, which seems to be faster (but not more effective) than Bayesian Optimization, CMA-ES and HBRKGA, which find their best at approximately 160 solutions for all datasets. The Fashion-MNIST and COSMOS datasets presented the fastest and slowest convergences, respectively.

\begin{table}[htb]
\centering
\caption{$p$-values resulting from applying the Wilcoxon test ($\alpha = 0.05$) to compare the baseline methods (Grid Search, Random Search, Bayesian Optimization, and CMA-ES) to HBRKGA.}
\label{tab:mww-results}

\begin{tabular}{@{}ccccccccc@{}}
\toprule
\multicolumn{1}{l}{} 
& \multicolumn{1}{c}{GS} 
& \multicolumn{1}{c}{RS}
& \multicolumn{1}{c}{BO}
& \multicolumn{1}{c}{CMA-ES}
\\ \midrule

\multirow{1}{*}{MNIST} & 0.00017  & 0.00068 & 0.00072 & 0.00015  \\ [0.5ex]
\multirow{1}{*}{MNIST-R} & 0.00017  & 0.01862 & 0.00016 & 0.00017 \\ [0.5ex]
\multirow{1}{*}{MNIST-IB} & 0.02323  & \textbf{0.31500} & \textbf{0.48100} & \textbf{0.90350}  \\ [0.5ex]
\multirow{1}{*}{MNIST-RotBack} & 0.00147  & 0.00021  & 0.00713 & \textbf{0.93230} \\ [0.5ex]
\multirow{1}{*}{MNIST-RandBack} &    \textbf{0.06352}    &    0.00018   &   \textbf{0.9370} & \textbf{0.4723} \\ [0.5ex]
\multirow{1}{*}{Fashion-MNIST}   & 0.00018 & 0.00018 & \textbf{0.08095} & \textbf{0.06954} \\ [0.5ex]
\multirow{1}{*}{Rectangles} & 0.00017 & 0.00026 & 0.00735 & 0.00638 \\ [0.5ex]
\multirow{1}{*}{COSMOS}   & 0.01709 & 0.01395 & 0.00357 & \textbf{0.14850} \\ [0.5ex] \hline
\end{tabular}%
\end{table}

We used the Wilcoxon non-parametric test~\citep{wilcoxon1992individual} to verify whether the samples are statistically significantly different from each other. We set the significance level $\alpha = 0.05$. The set of 10 runs of Grid Search, Random Search, Bayesian Optimization and CMA-ES were compared against the results of HBRKGA runs. The resulting $p$-values are presented in Table~\ref{tab:mww-results}. The cases in which there was no observed statistically significant difference between the distributions are highlighted in boldface. This occurred in some methods in the datasets MNIST-IB, MNIST-RotBack, MNIST-RandBack, Fashion-MNIST and COSMOS, especially with Bayesian Optimization and CMA-ES. For all the other methods and datasets, in the conditions studied, the results of HBRKGA are statistically significantly different than other methods.

\subsubsection{Computational Performance}
\label{sec:exp-results:comp:perf}

The computational performance results of each method are presented in Table \ref{tab:time-results}, which shows the mean execution time and their respective standard deviations for the ten experimental runs. The smallest results found are highlighted in each line. Random Search was able to outperform the other methods in 4 out of 8 datasets. Bayesian Optimization showed the highest processing time in all the experiments performed. It is also possible to highlight that HBRKGA surpassed other methods in the MNIST-RandBack and MNIST-RotBack datasets, which were the MNIST variants with the lowest result in the $F_1$ metric. The values of hyperparameters generated as a solution for each case directly influences the learning time of ANN. Higher learning rate values can make the model converge faster, while smaller values can make the learning time longer. Grid Search gets the best result for Fashion-MNIST and Rectangles datasets. 

\begin{table}[htb]
\centering
\caption{Average time results (in seconds) for 10 experimental runs. The best results are presented in bold face. The last line (labelled AVG) presents the average results considering all ten runs.}
\label{tab:time-results}
\resizebox{\columnwidth}{!}{%
\begin{tabular}{@{}cccccccccccccccc@{}}
\toprule
\multicolumn{1}{l}{} 
& \multicolumn{2}{c}{GS} 
& \multicolumn{2}{c}{RS}
& \multicolumn{2}{c}{BO}
& \multicolumn{2}{c}{CMA-ES}
& \multicolumn{2}{c}{HBRKGA} \\

\cmidrule(lr){2-3} \cmidrule(lr){4-5} \cmidrule(lr){6-7} \cmidrule(lr){8-9} \cmidrule(lr){10-11}
 
& AVG
& STD
& AVG
& STD
& AVG
& STD
& AVG
& STD
& AVG
& STD
\\ \midrule

\multirow{1}{*}{MNIST} & 11050 & 214 & \textbf{10836} & 303 & 32298 
                       & 884 & 13199 & 386   & 14474    & 474   \\ [0.5ex]
\multirow{1}{*}{MNIST-R}  & 13544 & 199 & \textbf{11213} & 288 & 28950
                          & 710 & 14323 & 308 & 15864 & 380\\[0.5ex]
\multirow{1}{*}{MNIST-IB} & 14284 & 120 & \textbf{13621} & 356 & 33020
                          & 1950 & 15666 & 250 & 19843 & 312\\[0.5ex]
\multirow{1}{*}{MNIST-RotBack} & 19823 & 286 & 19862 & 258 & 44985
                               & 2232 & 23425 & 569 & \textbf{19253} & 1807\\[0.5ex]
\multirow{1}{*}{MNIST-RandBack} & 25143 & 580 & 24540 & 856 & 49779
                                & 2232 & 25203 & 667 & \textbf{21988} & 513\\[0.5ex]
\multirow{1}{*}{Fashion-MNIST}   & \textbf{56811} & 126 & 59390 & 241 & 109471
                          & 10396 & 59560 & 847 & 57214 & 2094\\[0.5ex]
\multirow{1}{*}{Rectangles} & \textbf{2281} & 50 & 2357 & 98 & 9563
                            & 374 & 5563 & 185 & 6407 & 305\\[0.5ex]
\multirow{1}{*}{COSMOS}   & 2238 & 66 & \textbf{1157} & 167 & 8011
                          & 898 & 3658 & 401 & 3726 & 353\\[0.5ex] \hline
 \multirow{1}{*}{AVG}  & 18146 & 205 & \textbf{17872} &  230 & 39509 & 2459 & 20074 & 451 & 19846 & 779                    \\[0.5ex]
\end{tabular}%
}
\end{table}

\subsection{Ablation Study}
\label{sec:ablation}

In general, in an ablation study the goal is to understand the behavior of a system by removing/changing some components of it and observing the impact. In  this section, we describe an ablation study we conducted using the rectangle dataset (Section~\ref{sec:datasets}) to evaluate the impact of the Random-Walk component in the behavior of HBRKGA. The seed was fixed in HBRKGA(0) (without applying Random-Walk) and HBRKGA(3) (using Random-Walk with three steps) algorithms, where the solutions were sent to the generation and quality evaluation of the model in ANN without fixed seed. This process was repeated ten times in each algorithm.

The total number of solutions at the end of runs were kept the same in both algorithms, taking into account the adjustment by Random-Walk that increases the number of solutions in each generation. For HBRKGA(0), ten generations were used (the first reserved for a random initial solution) with a population size 24, generating 240 solutions in total. For the HBRKGA(3), the generations were reduced to 6 with a population size of 10, but adding 3 Random Walk steps in each population individual generated by the algorithm.

\begin{figure}
\begin{center}
\begin{tikzpicture}
      \begin{axis}[
      width  = 0.60*\textwidth,
      height = 8cm,
      major x tick style = transparent,
      ybar=2*\pgflinewidth,
      bar width=25pt,
      ymajorgrids = true,
      symbolic x coords={HBRKGA(0),HBRKGA(3)},
      xtick = data,
      scaled y ticks = false,
      enlarge x limits=0.50,
      ymin=0.9,
      legend cell align=left,
      legend style={at={(0.5,-0.12)},anchor=north},
  ]
      \addplot[style={fill=white},error bars/.cd, y dir=both, y explicit,
      error bar style={line width=1pt}]
          coordinates {
          (HBRKGA(0), 0.975) += (0,0.00195) -= (0,0.00195)
          (HBRKGA(3),0.98) += (0,0.00285) -= (0,0.00285)};
  \end{axis}
 \end{tikzpicture}
 \caption{HBRKGA behavior with (HBRKGA(3)) and without (HBRKGA(0)) the Random-Walk component. HBRKGA(3) reached the mean $F_1$ metric for 10 runs $0.98\pm 0.00285$ while HBRKGA(0) reached $0.975\pm 0.00195$.} \label{fig:brkgaXhbrkga}
 \end{center}
 \end{figure}
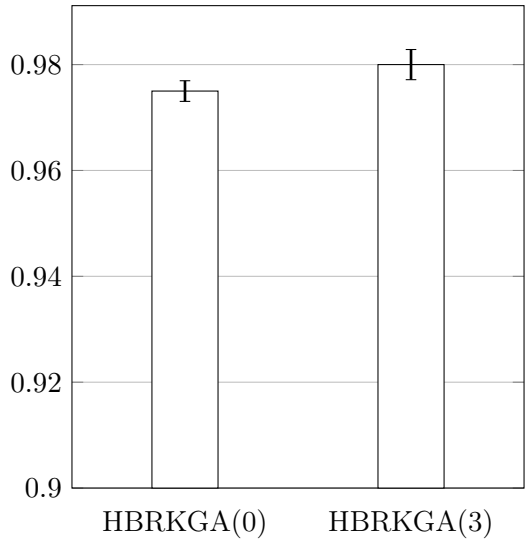

Figure~\ref{fig:brkgaXhbrkga} presents the results obtained in comparative experiments. The mean of $F_1$, inside each bar, indicates an advantage for HBRKGA(3), which increased the mean result obtained by HBRKGA(0) in the conditions studied. It is also possible to verify that HBRKGA(3) showed a more variance in $F_1$ than HBRKGA(0), however it maintained a better result taking into account the upper limit or the lower limit of the interval highlighted in red. This variations can be caused by ANN factors, like weights initialization and the hyperparameter value behavior generated by both algorithms.

\section{Conclusions}
\label{sec:conclusion}
In this paper we presented a new evolution-based approach to calibrate hyperparameters in a machine learning context. HBRKGA, the proposed hybrid method, combines a genetic algorithm (BRKGA) with a Random-Walk technique, with the goal of finding higher quality hyperparameter configurations. We performed several experiments in the context of artificial neural networks for solving classification problems. On the experiments conducted in this work comparing HBRKGA to other approaches (Grid Search, Random Search, Bayesian Optimization and CMA-ES), HBRKGA produced better average $F_1$ values (measured in a separate validation dataset) in 6 out of 8 datasets. By using the Wilcoxon test, we found HBRKGA to be statistically significantly better than the other methods in three datasets, while being highly competitive in the other datasets.

We also performed an ablation study to assess the impact of the Random-Walk component. In particular, we compared the full-blown variant of HBRKGA with the one in which we removed the Random-Walk component. It was possible to observe a statistically significant difference in the average $F_1$ values, with the same experimental conditions for both variants. We conclude that the Random-Walk component, albeit simple to implement, is a crucial part of HBRKGA, allowing it to inherit nice properties of Random Search already identified in previous work~\citep{bergstra2012random}.

For future work, we will investigate the application of HBRKGA to other machine learning methods and tasks. Besides, we will investigate its use in tuning hyperparameters of particular ANN architectures, such convolutional neural networks. We will also investigate new alternatives to perform the perturbation in HBRKGA instead of Random-Walk; one possibility is to add the points generated and evaluated in HBRKGA to Bayesian Optimization, using these previously known solutions for new acquisitions during the Gaussian Process.

\setcounter{magicrownumbers}{0}

\bibliographystyle{elsarticle-harv}

\section*{Computer Code Availability}
We implemented the neural network models presented in the experiments of this paper using \textit{Tensorflow} 1.13, an open-source Deep Learning framework. All of the source code used in the validation experiments is publicly available at \url{https://github.com/MLRG-CEFET-RJ/HBRKGA}.

\section*{Data Availability}
All datasets used in the experiments of this paper (MNIST and variations, rectangles, COSMOS and Fashion-MNIST) can be downloaded at \url{https://doi.org/10.5281/zenodo.4252922}, an open-source online data repository.

\section*{Acknowledgment}

The authors thank CNPq, CAPES, FAPERJ, and CEFET/RJ for partially funding this research.

\bibliography{references}

\end{document}